%% file: root.tex
\documentclass[a4paper]{svproc}

\usepackage{url}

\usepackage{graphicx}
\usepackage{multirow}
\usepackage{footmisc}
\usepackage{natbib}
\usepackage{todonotes}
\usepackage{siunitx}
\usepackage{dsfont}
\usepackage{hyperref}
\usepackage{mathtools}
\usepackage{amssymb}
\usepackage{tabularx}
\usepackage{pifont}%
\newcommand{\cmark}{\ding{51}}%
\newcommand{\xmark}{\ding{55}}%

\DeclarePairedDelimiter\set\{\}

\begin{document}
\mainmatter              %
\title{A Billion Ways to Grasp:\\An Evaluation of Grasp Sampling Schemes\\on a Dense, Physics-based Grasp Data Set}
\titlerunning{A Billion Ways to Grasp}  %
\author{Clemens Eppner \and Arsalan Mousavian \and Dieter Fox}
\authorrunning{Eppner et al.} %
\institute{NVIDIA USA,\\
\email{ceppner@nvidia.com}%
}

\maketitle              %

\begin{abstract}
Robot grasping is often formulated as a learning problem. With the increasing speed and quality of physics simulations, generating large-scale grasping data sets that feed learning algorithms is becoming more and more popular. An often overlooked question is how to generate the grasps that make up these data sets. In this paper, we review, classify, and compare different grasp sampling strategies. Our evaluation is based on a fine-grained discretization of SE(3) and uses physics-based simulation to evaluate the quality and robustness of the corresponding parallel-jaw grasps. Specifically, we consider more than 1~billion grasps for each of the 21~objects from the YCB data set. This dense data set lets us evaluate existing sampling schemes w.r.t.~their bias and efficiency. Our experiments show that some popular sampling schemes contain significant bias and do not cover all possible ways an object can be grasped.
The data is available at~\url{https://sites.google.com/view/abillionwaystograsp/}.
\end{abstract}
\input{00_introduction}

\input{01_related_work}

\input{03_taxonomy}
\input{05_evaluation}

\input{07_experiments}
\input{08_discussion}

\input{09_conclusions}

\bibliographystyle{spbasic}
\bibliography{references}

\end{document}

%% file: 00_introduction.tex
\section{Introduction}

Grasping is a fundamental skill for any robotic manipulation system. Most commonly it is solved in a data-driven fashion~\citep{bohg2013data}, either through supervision~\citep{Mahler2017} or reinforcement~\citep{Levine2016,kalashnikov2018qt}. To satisfy the data hunger of these learning methods, grasps are often labeled in simulation.

The advantages of evaluating grasps in simulation are manifold: data collection can be scaled easily, grasp conditions can be controlled, robots won't break, resetting is trivial, and the supervision signal benefits from a fully observable environment.
Although the gap between simulation and reality needs to be addressed, it has been shown that models trained exclusively with synthetic grasp data can perform successfully in the real world~\citep{Mahler2017}.

Generating synthetic grasp data is usually based on heuristics that select a gripper pose relative to the object. Oftentimes these heuristics don't get much attention in grasp learning publications and occupy only a minor paragraph.
In this paper, we thoroughly analyze and compare these different sampling schemes.
We focus our evaluation on quantifying the grasp coverage for each heuristic.
Do some sampling schemes cover the space of all possible grasps of an object better than others?

To answer this question empirically we need a ground truth grasp density for a given object.
We acquire this ground truth by discretizing the space of all possible grasps with high resolution and evaluate them in a physics simulation~\citep{macklin2014unified}.
This reference data set contains dense grasp sets for 21~objects of the YCB~object set~\citep{calli2017yale}.
To the best of our knowledge there has been no prior attempt to exhaustively describe all possible grasps for an object.

Note, that our analysis is not limited by the realism of the results produced in the physics simulation.
However, we show that the simulated grasps can be successfully executed on a real robotic system~(Sec.~\ref{sec:real robot}). 
Furthermore, due to the denseness of the data we can generate robust versions of the original grasp sets and use those for evaluation.

\paragraph{Contribution}
Our contribution is two-fold:
\begin{enumerate}
    \item We present a set of all possible parallel-jaw grasps for 21 objects. It is generated by discretizing SE(3) with a resolution of (\SI{5}{\mm}, \SI{7.5}{\degree}) and executing more than a billion grasps in a physics simulator.
    \item We use this data set to study and compare  different sampling schemes for grasping. Our comparison shows how grasp coverage is affected by using different samplers. Based on these results we recommend sampling strategies for generating large-scale grasp data sets.
\end{enumerate}

\paragraph{Organization}
The paper is structured as follows. First, we review existing grasp sampling schemes that are used for generating data sets. We then categorize those methods into a coherent taxonomy, and present our evaluation criteria. Finally, we compare a number of sampling schemes and discuss their pros and cons.

%% file: 01_related_work.tex
\section{Related Work}
\label{sec:related work}
To the best of our knowledge, there has been no prior evaluation of different sampling strategies for generating grasp data sets.
In contrast, the field of sampling-based motion planning~\citep{lavalle2006planning} is rife with sampling strategies based on heuristics. However, it has been show that no sampling strategy outperforms all others in all scenarios~\citep{lindemann2005current,elbanhawi2014sampling}.

In the following we briefly review grasp sampling strategies, clustered according to their use case.
Since our evaluation uses a physics simulator, we also give a brief overview about the usage of physics simulation for robot grasping. 

\subsubsection{Grasp Sampling for Planning/Inference}
\label{subsec:sampling for planning}
Sampling-based techniques are often used to find optimal grasps given an analytical or learned model of grasp quality. Examples are simulated annealing~\citep{ciocarlie2007dimensionality,hang2016hierarchical} or the cross-entropy method~\citep{Mahler2017,yan2017learning}.
We do not include these black-box optimization methods in our analysis for two reasons. First, these methods are used for finding an optimum, while we are interested in discovering the entire grasp distribution. Second, since these sampling approaches require models that can be evaluated quickly, their performance also depends on the quality of the model approximation.

\subsubsection{Grasp Sampling for Generating Real-World Data Sets}
The Cornell grasping dataset~\cite{jiang2011efficient} contains $\approx5K$~human-labelled grasps for 280~objects which are represented as rectangles in the image plane. Since the data is relatively sparse~($\approx18$~grasps per object), human-sampled, and view-points correlate with the object's equilibrium poses, bias is inevitable.
\cite{pinto2016supersizing} collect $50K$~top-down grasps autonomously by sampling random grasp points and orientations in $SE(2)$. Again, the sampling method is biased since sampling only happens in planes parallel to the object's equilibrium poses.

\subsubsection{Grasp Sampling for Generating Synthetic Datasets}
Our main focus is on sampling methods that are used to generate large-scale synthetic datasets via physics simulation.
The Columbia grasp database~\citep{goldfeder2008columbia} generated grasps using the simulated annealing-based Eigengrasp planner~\citep{ciocarlie2007dimensionality} described above.
 
\cite{zhou20176dof}~learn from grasping data generated in a physics simulation. Their sampling scheme samples random lines that intersect the object's center of mass. The hand pose is then determined by shifting along this line in an arbitrary orientation. This scheme ensures that the space near the object's COM is more densily sampled than poses that are further away.

\cite{Kappler2015}~present a large dataset of grasps generated by the sampling scheme of~\cite{diankov2010automated}.
It samples hand approach vectors close to the object's surface normals and chooses random roll angles and standoff distances.
In a similar vein, the grasp dataset by \cite{kleinhans2015g3db} uses surface normals to sample the hand position while the orientation is chosen randomly. 
\cite{veres2017integrated} sample grasp poses around the object such that the hand's approach vector intersects the object's bounding box.  
In the following section~(Sec.~\ref{sec:taxonomy}) we will group all these sampling methods as approach-based schemes.

In contrast to approach-based samplers are antipodal-based sampling schemes.
They sample directly the potential contact points with the object.
Examples are the data sets used in \citep{Mahler2017,ten2018using,liang2018pointnetgpd}.

\subsubsection{Physics Simulation for Grasping}
Our analysis relies on the evaluation of grasp candidates in a physics simulation~\citep{macklin2014unified}.
Simulating stable grasps is challenging~\citep{erez2015simulation} and the results never fully transfer to the real world~\citep{collins2018quantifying}.
However, it has been shown that simulation provides significantly more information about grasp success than traditional grasp quality metrics, such as force-closure analysis~\cite{kim2013physically}.
This is due to focusing on the entire grasp process including object dynamics, instead of only measuring the quality of the established contacts.

%% file: 03_taxonomy.tex
\section{A Taxonomy of Sampling Strategies for Grasping}
\label{sec:taxonomy}
We define a grasp as a combination of a pre-grasp and a closing motion. The pre-grasp~$g \in SE(3)\times\mathds{R}^n$ describes the pose and configuration of the hand~($n$ is the number of internal DoF) prior to the execution of a controller that represents the closing motion. In the remainder we assume that the closing controller position-controls the hand pose and uses a force-based control law to close the fingers.
Given a fixed closing controller, we  we will focus our evaluation on parallel-jaw grippers~($n = 1$), i.e., the space of all possible grasps is $SE(3)\times\mathds{R}$. We assume the fingers to be maximally opened during the pre-grasp which further reduces the grasp space to~$SE(3)$. 

\begin{table}[t]
    \centering
    \begin{tabularx}{\textwidth}{|c|c|X|p{5cm}|}
    \hline
    \multicolumn{2}{|c|}{Guided by} & \multirow{3}{*}{Category} & \multirow{2}{*}{Publications generating} \\
    Grasp & Object & & \multirow{2}{*}{Grasp Data Sets} \\
    Result & Geometry & & \\\hline
    \multirow{6}{*}{\xmark} & \multirow{2}{*}{\xmark} & Uniform &  \\\cline{3-4}
     &  & Non-uniform & \cite{zhou20176dof} \\\cline{2-4}
     & \multirow{4}{*}{\cmark} & \multirow{2}{*}{Approach-based} & \cite{Kappler2015,kleinhans2015g3db,veres2017integrated} \\\cline{3-4}
     & & \multirow{2}{*}{Antipodal-based} & \cite{Mahler2017,ten2018using}\\
    \hline
    \cmark & \cmark & Adaptive & \cite{goldfeder2008columbia}\\\hline
    \end{tabularx}
    \caption{A taxonomy of grasp samplers. See text for detailed explanation.}
    \label{tab:taxonomy}
\end{table}

We compare different sampling methods of this grasp space regarding to how well they cover all successful grasps. In the previous section we reviewed those samplers according to their application scenario, i.e. whether they where used to generate data or for planning. Now, we want to subdivide them in more detail into categories based on what information they use and how they parameterize the grasp space.
The taxonomy shown in Tab.~\ref{tab:taxonomy} classifies grasp sampling along the following criteria:

\subsubsection{Guided by Grasp Result}
Our fist broad distinction is whether a grasp sampler evaluates the grasp quality function and uses this outcome when drawing subsequent samples.
This is independent of the actual realization of the grasp quality function. It could be any classical grasp metric~\citep{roa2015grasp}, a physics simulation, or even the physical execution of the grasp on a real platform.

We focus our empirical evaluation on sampling methods that are not guided by this information. Since most grasp quality functions depend on contact and are noncontinuous, grasp information is very local which is of limited value when generating datasets that contain diverse grasps that should fully cover an object. 
The large majority of existing grasp datasets is generated by this methods that are not guided by the grasp outcome.

\subsubsection{Guided by Object Geometry}
Most grasp sampling methods are guided by surface information of the object. This is often done by parameterizing the grasp using surface normals, either by aligning the hand's approach vector (or the palm's surface normal) with the object's surface normal or by aligning the expected finger contact normals with object's surface.
We will show in our empirical analysis that although these methods are effective at generating grasps, they are biased, i.e., the resulting grasps do not fully cover all possible grasps of an object. 

\subsubsection{Uniform Samplers}
Without using any geometric information about the object or the outcome of a grasp sample the best thing we can do is sampling the bounded $SE(3)$ space uniformly.
The uniformity of samples can be expressed by measures like discrepancy and dispersion~\citep{lavalle2006planning}. As a result multiple sequences have been proposed that result in better uniformity than those produced by pseudo-random number generators.
Among low-discrepancy sampling there are three categories: Halton~\citep{halton1960efficiency}/Hammersley sequences, (t,s)-sequences and (t,m,s)-nets, and lattices such as the Sukharev grid~\citep{sukharev1971optimal}.
Lattices are finite point sets which limits their applicability. But incremental grids for $SO(3)$~\citep{yershova2010generating} and $SE(3)$~\citep{lindemann2004incremental} have been proposed.

Note, that low-discrepancy sampling techniques are not limited to uniform sampling schemes. All of the following sampling methods can benefit from applying low-discrepancy sampling for their parameters or subsets of them. But care needs to be taken, given that low discrepancy in parameter space not necessarily leads to low discrepancy in $SE(3)$.
In our evaluation we include a uniform sampling scheme based on a pseudo-random number generator.

\subsubsection{Non-uniform Samplers}
There are only few sampling methods that do not exploit information about the object's geometry but still sample non-uniformly.
One example is the approach taken by~\cite{zhou20176dof}. They sample random lines that go through the origin (i.e. center-of-mass) of the object, with the directions being distributed uniformly. Evenly spaced points are chosen along a line that form the translation of the grasp. The orientation is sampled randomly.  This scheme results in a higher density of grasp samples closer to the COM of the object.

\begin{figure}[t]
    \centering
    \includegraphics[width=.5\textwidth]{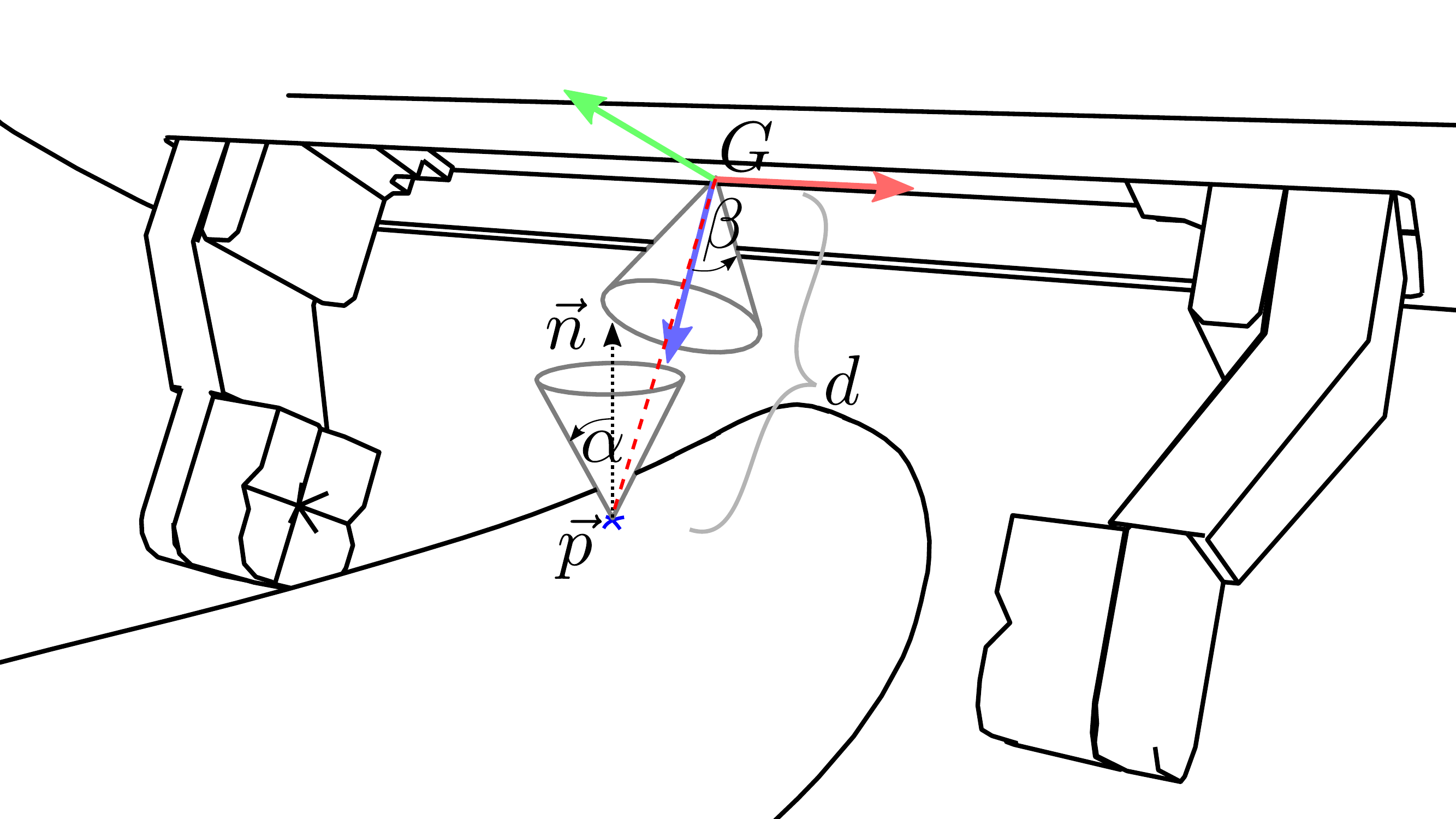}%
    \includegraphics[width=.5\textwidth]{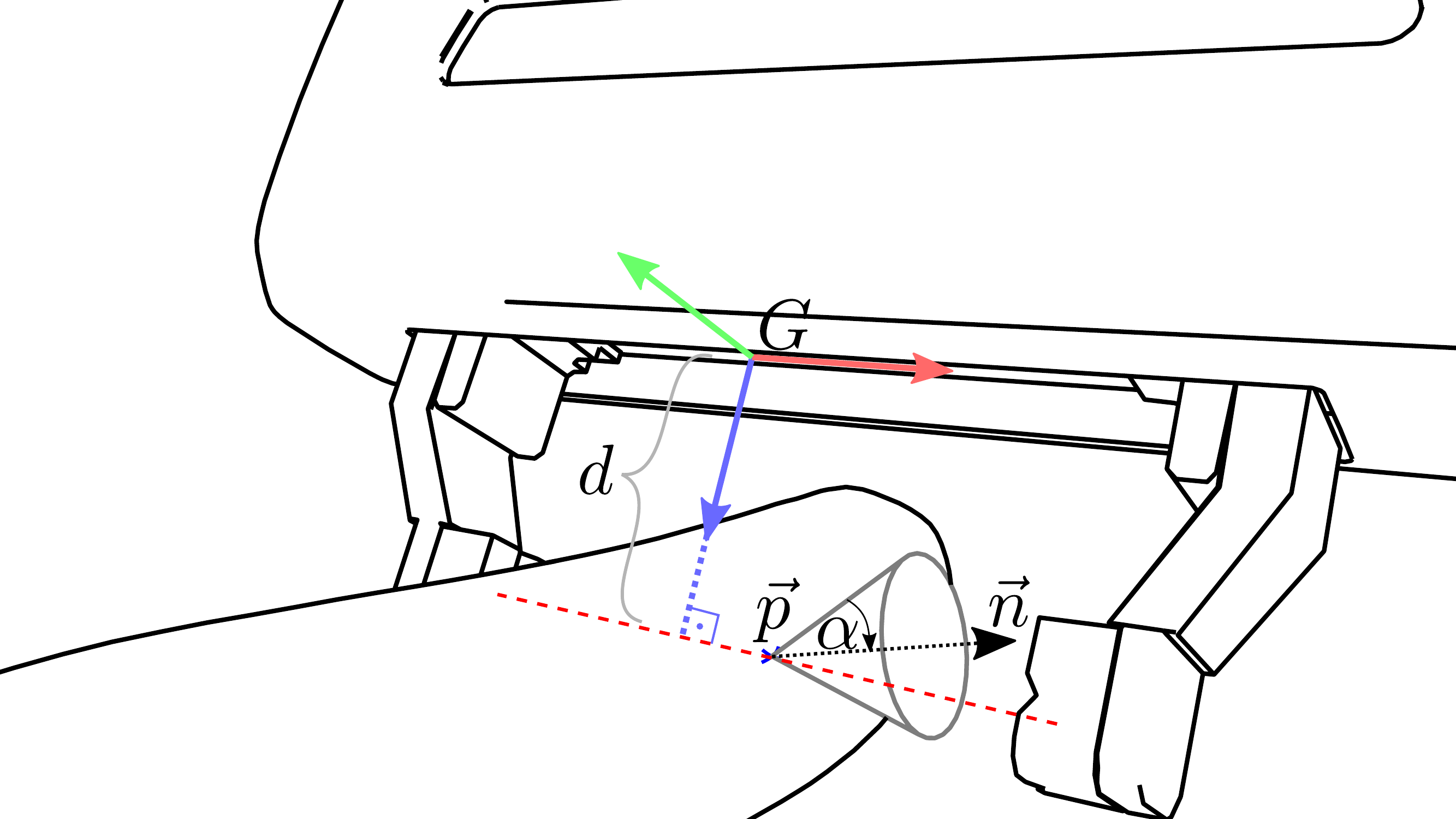}%
    \caption{Parameterization of the approach-based grasp sampling schemes~(\emph{left}) and the antipodal-based schemes~(\emph{right}). See text for details.}
    \label{fig:sampling schemes}
\end{figure}

\subsubsection{Approach-based Samplers}
The majority of grasp data sets are generated via approach-based sampling methods. The approach vector of a gripper is the direction in which the grasp pose is approached and usually aligns with the palm's surface normal.
The sampling scheme most commonly aligns the approach vector with the surface normal of a randomly sampled point on the object. But there are a number of variants among those techniques.
Points on the object surface are either sampled uniformly or selected by casting rays from a bounding box~\citep{diankov2010automated, Kappler2015}.
Another approach uses the surface points and normals of a fitted primitive~(box, sphere, cylinder) to sample grasps~\citep{miller2003automatic}.
\cite{veres2017integrated} also sample the approach vector of the gripper.

For evaluation, we parameterize the most important subgroup of approach-based sampling methods as follows: Given a point on the object's surface and its corresponding normal, a direction is chosen whose angular difference with the normal is below $\alpha$, a standoff is chosen between zero and the length of the fingers, and an approach vector is chosen whose angular difference with the chosen direction is below~$\beta$~(see Fig.~\ref{fig:sampling schemes}). The hand's roll around the approach vector is finally chosen to be between $0$ and $2\pi$.
Our evaluation contains strategies for the following~$(\alpha, \beta)$: $(0,0)$, $(0,\pi)$, , and $(\frac{\pi}{2},0)$.

\subsubsection{Antipodal-based Samplers}
In contrast to the approach-based heuristics, another popular group of methods tries to sample directly in the space of possible contact points between object and hand. In addition, these methods exploit the conditions under which antipodal grasps create force-closure~\citep{Mahler2017,ten2018using}.

In contrast to approach-based samplers, it is non-trivial to scale antipodal-based samplers to multi-fingered hands and beyond antipodal grasps. This is due to the fact that there is no bijective mapping between hand configuration and contact locations.

For evaluation, we parameterize the antipodal-based sampling strategies as follows: Given a point on the object's surface surface and its corresponding normal, an antipodal point is chosen by finding the farthest location of intersection with the object along a ray whose angular difference with the normal is below~$\alpha$. Given the two antipodal contact points, the gripper pose is defined by choosing the center point along the ray, a rotation around the ray between $0$ and $2\pi$, and a standoff in the interval $[s_{\text{min}}, 0]$~(see Fig.~\ref{fig:sampling schemes}).
Our evaluation contains strategies for the following $(\alpha, s_{\text{min}})$: $(\frac{\pi}{6}, 0)$ and $(\frac{\pi}{2}, 0)$.

\subsubsection{Adaptive Samplers}
We group all strategies that select new samples based on the outcome of previous grasp samples into the category of adaptive samplers.
All planning approaches described in the related work are adaptive samplers~(Sec.~\ref{subsec:sampling for planning}).
This includes methods like simulated annealing~\citep{ciocarlie2007dimensionality}, cross-entropy method, importance sampling, or Bayesian optimization.
We do not include any of those methods in our evaluation.

%% file: 05_evaluation.tex
\section{Evaluation}
\label{sec:evaluation}

\subsection{Grasp Evaluation Metrics}
Our evaluation metrics are based on distances between grasps.
Similar to \cite{mahler2016privacy} we use a weighted metric. Let $g, h \in SE(3)$ be two grasps, with $g_p, h_p \in \mathds{R}^3$ being their positions and $g_q, h_q \in \mathcal{S}^3$ their orientations represented as unit quaternions. The distance between $g$ and $h$ is defined as:
\[
\rho (g, h) = \omega \|g_t - h_t \|_2 + \arccos (|\langle g_q ,  h_q \rangle|),
\]
where $\omega \in \mathds{R}$ is a weight that relates rotation and translation. Unlike~\cite{mahler2016privacy} we do not select it depending on the size of the object. Instead we keep it constant, such that a pure translation of~\SI{1}{\mm} equals a pure rotation of~\SI{1}{\degree}.
Given the distance metric $\rho$, we now show which performance metrics we use to compare the different sampling mechanisms.

\subsubsection{Grasp Coverage}
Our main objective is to find sampling methods that capture the reference grasp distribution of an object.
We define different measures of grasp coverage that capture different properties as follows.
The set $\mathcal{X}$ contains all grasps sampled by a particular method, while $\mathcal{R}$ is the reference set of all successful grasps found in simulation.
Our first metric is defined as:
\[
\text{cov}_1(\mathcal{X}, \mathcal{R})_{\epsilon} = \frac{| \{g \mid g \in \mathcal{R} \wedge \exists x \in \mathcal{X}: \rho(g, x) \leq \epsilon \}|}{|\mathcal{R}|},
\]
where $\epsilon \in \mathds{R}$ defines the maximum distance that two grasps are considered equal.
Although $\text{cov}_1$ is intuitive it is sensitive to the choice of~$\epsilon$. It can even happen that the ordering of sampling methods according to $\text{cov}_1$ changes with different choices of $\epsilon$.

To circumvent this problem we also report a grasp coverage measure based on dispersion. This metric was used in~\citep{mahler2016privacy} and is defined as follows: 
\[
\text{cov}_2(\mathcal{X}, \mathcal{R}) = \exp\left(-\max_{g \in \mathcal{R}} \min_{x \in \mathcal{X}} \rho(x, g)\right)
\]
Since $\text{cov}_2$ is the longest of all shortest paths between $\mathcal{X}$ and $\mathcal{R}$, it can be dominated by outliers in $\mathcal{R}$. This is possible because the reference set is generated in a physics simulation. To get a more representative coverage measure we also report the average over all shortest paths: 
\[
\text{cov}_3(\mathcal{X}, \mathcal{R}) = \exp\left(-\frac{1}{|\mathcal{R}|}\sum_{g \in \mathcal{R}} \min_{x \in \mathcal{X}} \rho(x, g)\right)
\]

Note, that the computational bottleneck of all coverage calculations is the nearest-neighbor search, especially since we are dealing with large sets of up to millions of elements. In our implementation we use the SE(3)~k-d~tree by~\cite{ichnowski2015fast}.

\subsubsection{Precision}
Oftentimes learning approaches for grasping are based on a critic or discriminative model that predicts the quality of a given grasp. Training data for such models needs to be balanced, i.e., it should roughly contain as much positive as negative grasps. Since this is not captured by the coverage metrics, we also evaluate the different sampling schemes w.r.t. their precision. Precision is defined by the ratio of successful grasps among all sampled ones.

\subsection{Grasp Robustness}
\label{subsec:robust grasps}
Grasp success is very sensitive to the accurate reproduction of the contact configuration between hand and object. Slight variations in the positioning of the hand can lead to vastly different outcomes. Grasp planning approaches have addressed this by incorporating noise models for computing grasp quality metrics~\citep{weisz2012pose} or in physics simulations~\citep{kim2013physically}.

Similarly, we define the robustness of a grasp as the portion of successful grasps in its $\epsilon$-neighborhood.
Given a grasp $g \in \mathcal{R}$, a grasp set $\mathcal{R}$ with a constant grasp density, and an indicator function $\mathbf{1}_{succ}$ denoting a successful grasp, we define:
\[
\text{robust}_{\epsilon}(g) =  \frac{|\{g \in \mathcal{R}: \rho(g, x) \leq \epsilon, \mathbf{1}_{succ}(g) = 1\}|}{|\{g \in \mathcal{R}: \rho(g, x) \leq \epsilon|}
\]
Consequently, the robust version of a grasp set~$\mathcal{R}$ is defined as:
\[
\text{robust}_{\epsilon, \gamma}(\mathcal{R}) = \set[\bigg]{ g \mid g \in \mathcal{R},\; \text{robust}_{\epsilon}(g) \geq \gamma },
\]
where $\gamma \in [0, 1]$ is the robustness threshold.
In our evaluation we will also report the performance of the different grasp samplers w.r.t. the robust coverage metrics:
\[
\text{cov}_i^{\epsilon, \gamma}(\mathcal{X}, \mathcal{R}) =  \text{cov}_i(\mathcal{X},\; \text{robust}_{\epsilon, \gamma}(\mathcal{R})).
\]

\subsection{Evaluation in Simulation}

\begin{figure}[t]
    \centering
    \includegraphics[width=0.33\textwidth]{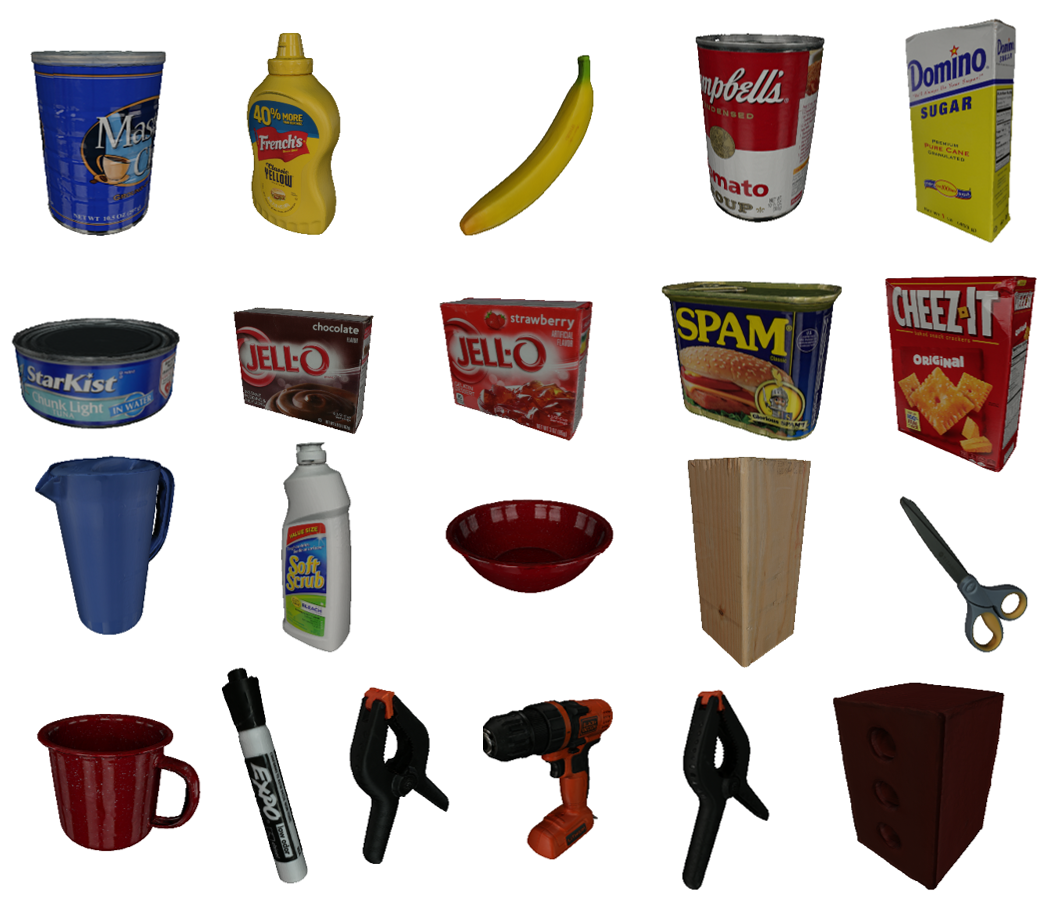}\hspace{1cm}%
    \includegraphics[width=0.5\textwidth]{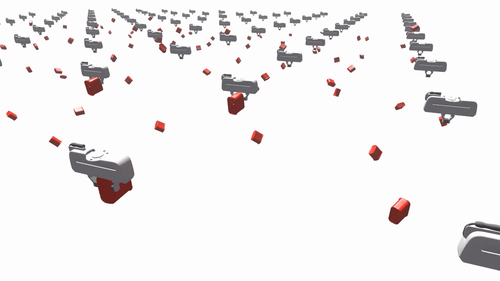}%
    \caption{Objects from the YCB~dataset~(left) were grasped in simulation by a parallel-jaw gripper~(right).}
    \label{fig:objects_and_sim}
\end{figure}

We evaluate grasps in a physics simulation. This allows us to scale our evaluation to extremely large quantities of grasp attempts (billions, in contrast to hundreds of thousands in real-world setups~\citep{Levine2016}). It also allows us to control all aspects of the data collection process, generating dense grasp distributions for single objects. We use the physics simulator FleX~\citep{macklin2014unified} and 21 object meshes of the YCB dataset~\citep{calli2017yale}, shown in Fig.~\ref{fig:objects_and_sim}. We assume a constant friction coefficient of~$1.0$ between the rubber pads of the Franka Panda gripper and all objects. All objects are assumed to have a constant density. The grasps are simulated in free space, without any gravity applied~(similar to~\cite{zhou20176dof}). Given an initial hand position, the gripper closes its fingers (using a force-based control scheme) and executes a pre-defined motion trajectory that involves linear shaking along the approach vector and angular shaking around the finger closing direction.
We record the amount of motion the object undergoes during finger closing and shaking. We also record whether the objects stays between the fingers until the end of the simulation.

Note, that our analysis is not limited to the evaluation in a physics simulator. The sampling schemes could also be evaluated against a number of classical grasp quality metrics~\citep{roa2015grasp}.
But given the evidence that classical metrics are very sensitive w.r.t. contact point locations and do not capture stability, we think simulation is a more realistic way to evaluate grasps. The experimental sections provides supporting evidence that the data generated in simulation is transferable to the real world.

%% file: 07_experiments.tex
\section{Experimental Results}

\subsection{Physics-based Reference Data}
We simulated the grasp outcomes for 21~different objects from the~YCB~dataset~\citep{calli2017yale}. Grasps are evenly spaced on a grid in~$SE(3)$ with \SI{5}{\mm} between hand positions, and \SI{7.5}{\degree} between neighboring orientations. Evenly distributed orientations were ensured by applying the method of~\cite{yershova2010generating}.
The simulation was done in FleX~\citep{macklin2014unified} using a model of the 1-DOF Franka Panda gripper.

\begin{figure}[t]
    \centering
    \includegraphics[height=3cm]{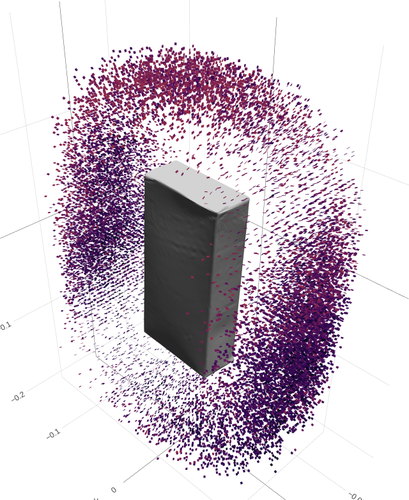}\hfill%
    \includegraphics[height=3cm]{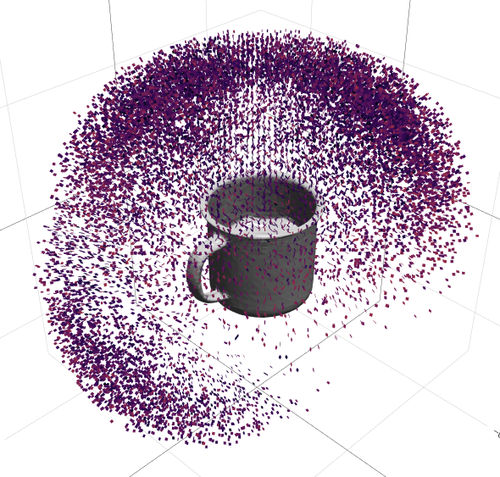}\hfill%
    \includegraphics[height=3cm]{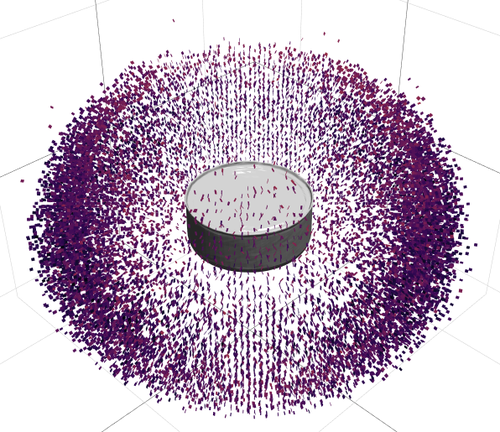}\hfill%
    \includegraphics[height=3cm]{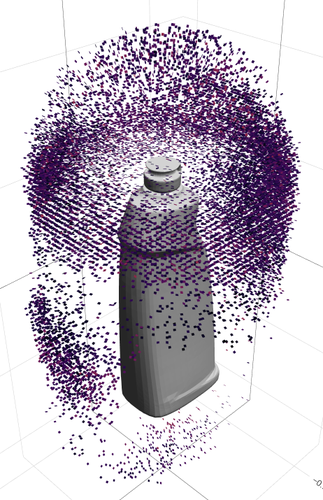}\\%
    \raisebox{-0.5\height}{\includegraphics[width=2cm]{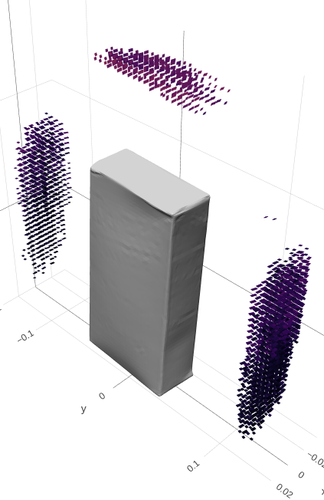}}\hfill%
    \raisebox{-0.5\height}{\includegraphics[width=3cm]{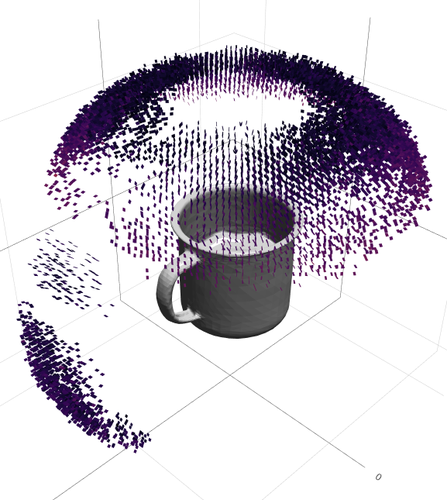}}\hfill%
    \raisebox{-0.5\height}{\includegraphics[width=2.5cm]{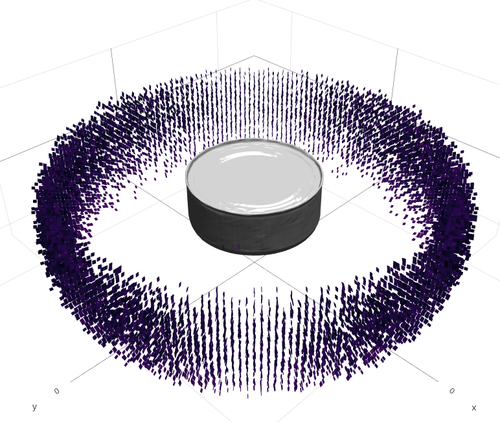}}\hfill%
    \raisebox{-0.5\height}{\includegraphics[width=2cm]{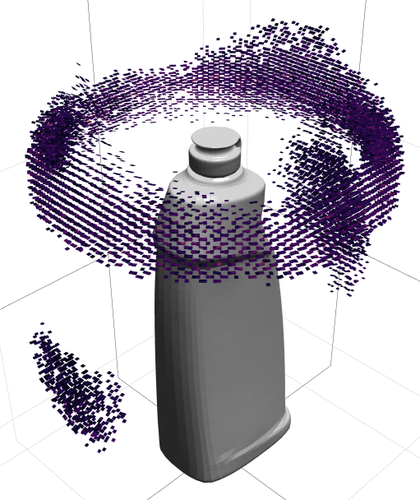}}\\%
    \caption{Four example objects~(left to right: sugar box, mug, tuna fish can, bleach cleanser) and the resulting successfully simulated grasps. Each colored point indicates a successful grasp pose. The bottom row shows robust versions of the grasp sets.}
    \label{fig:example_gt_densities}
\end{figure}

For each object, we simulated only those grasps that passed a collision test and which had a nonempty object volume between the fingers. All other grasps were marked as failures. In total, $\approx317$~billion grasps were sampled, of which~$\approx1$~billion passed~(\SI{0.32}{\%}) the tests and were simulated in FleX. Simulations were run on 100~GPUs for one and a half months. We simulated 225 grasps in parallel on a single GPU, which lasted \SI{90}{s} on average.
Out of all grasps $\approx156$~million were successful~(\SI{15.57}{\%}).
Fig.~\ref{fig:example_gt_densities} shows successful grasps for a few objects.

\subsection{Real-World Robot Experiments}
\label{sec:real robot}
To verify the simulated reference grasps, we conducted experiments in the real world with a 7-DOF Franka Panda manipulator equipped with a 1-DOF parallel-jaw gripper.
Since the grasps are defined in object coordinates, we need to estimate the object poses.
We use state-of-the-art object pose detectors~PoseCNN~\citep{xiang2017posecnn} and DeepIM~\citep{li2017deepim} to get an initial estimate and further refine it with DART~\citep{schmidt2014dart} using depth.

\begin{figure}[t]
    \centering
    \begin{tabular}{ccccc}
         \includegraphics[width=0.19\textwidth]{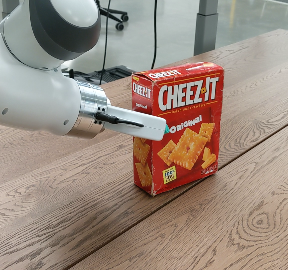} & 
         \includegraphics[width=0.19\textwidth]{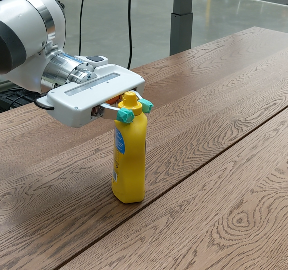} &
         \includegraphics[width=0.19\textwidth]{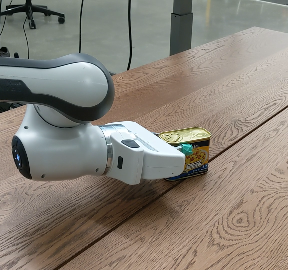} &
         \includegraphics[width=0.19\textwidth]{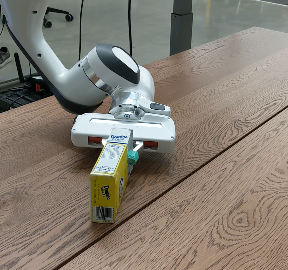} & 
         \includegraphics[width=0.19\textwidth]{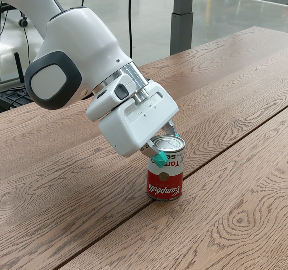}%
    \end{tabular}
    \caption{Example grasps from the simulated data set executed on the real robot.}
    \label{fig:real}
\end{figure}

Since it is impossible to evaluate all the reference grasps with the real robot, we verified a subset of grasps on five objects that are shown in~Fig.~\ref{fig:real}. For each object, five diverse grasps are chosen and executed. Success of each grasp in these experiments depends on the accuracy of the estimated object pose, control error, and also the quality of reference grasp. For each object, 100~grasps are sampled from the robust set of grasps for each object using farthest point sampling. The grasps that lead to collisions with the support surface are removed. From the remaining grasps, five diverse grasps are chosen to be executed. Out of the 25 grasps only three failed. This is due to the discrepancy between real-world physics and simulation. For example, objects have uniform density in the simulator, they are completely rigid and also exhibit different friction coefficients.
For a video of the experiments, see~\url{https://bit.ly/2HWEI2r}.

\subsection{Comparison of Different Sampling Methods}
We compared the different grasp sampling methods presented in Sec.~\ref{sec:taxonomy}. We ran each sampling method on all objects and calculated the different evaluation metrics presented in Sec.~\ref{sec:evaluation}.
We assume that a grasp pose that is in collision with the object is invalid as well as a grasp pose whose volume between the gripper's fingers does not intersect with any part of the object.
For all evaluated methods, we reject samples that do not pass these two tests.

\subsubsection{Grasp Coverage}
\begin{figure}[t]
    \centering
    \includegraphics[width=0.7\textwidth]{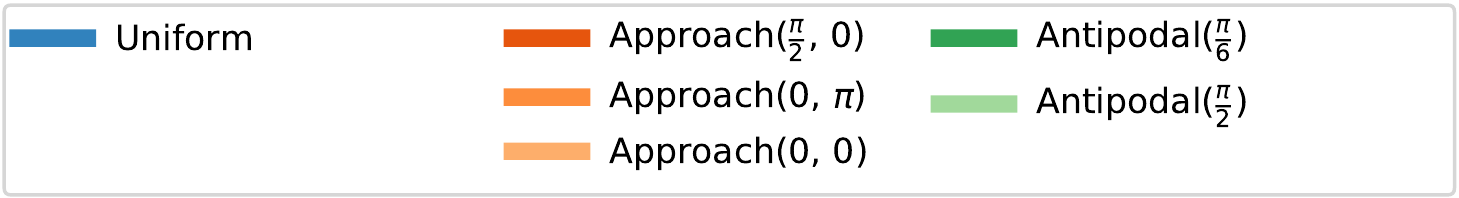}\\
    \includegraphics[width=0.5\textwidth]{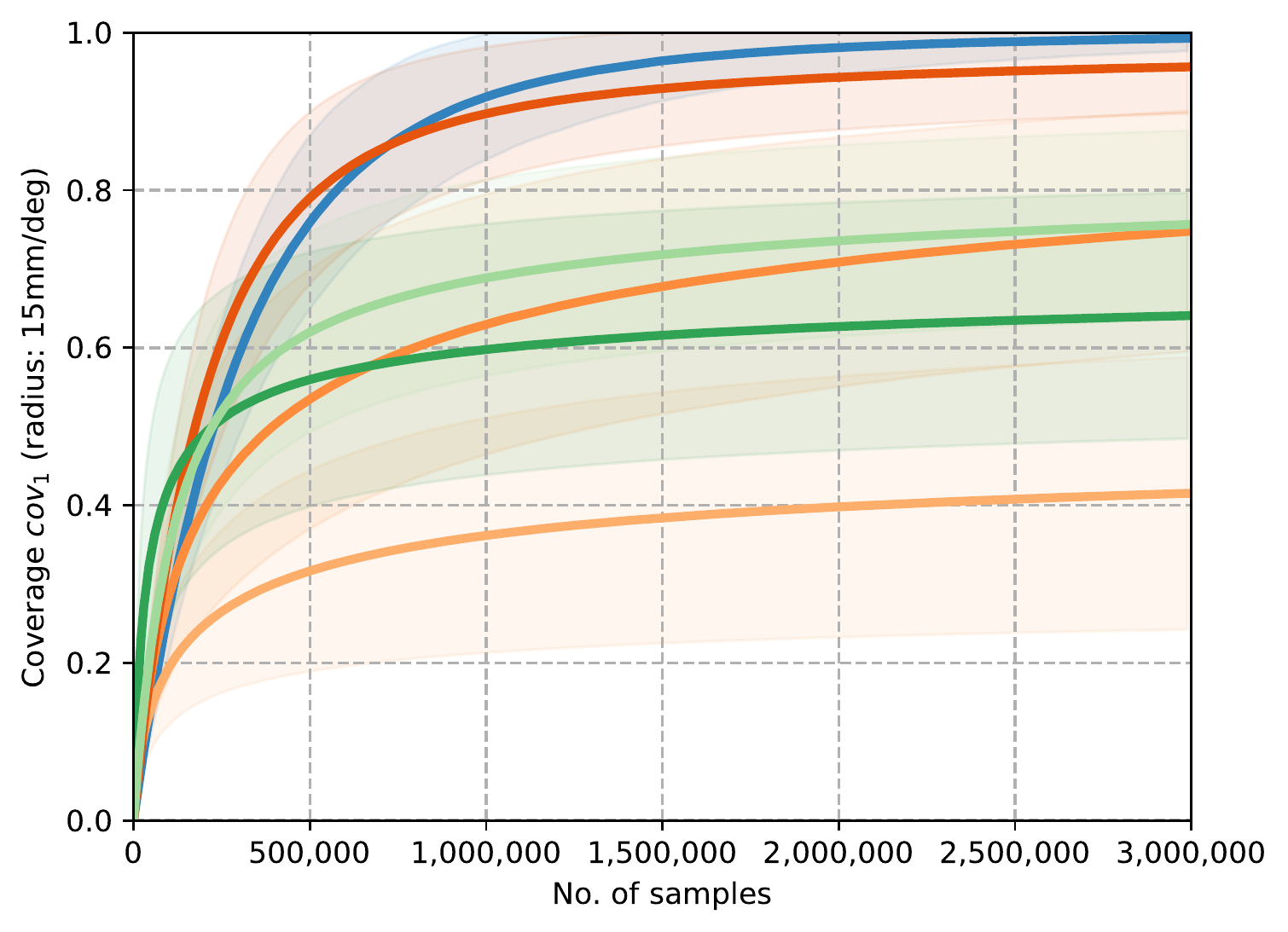}%
    \includegraphics[width=0.5\textwidth]{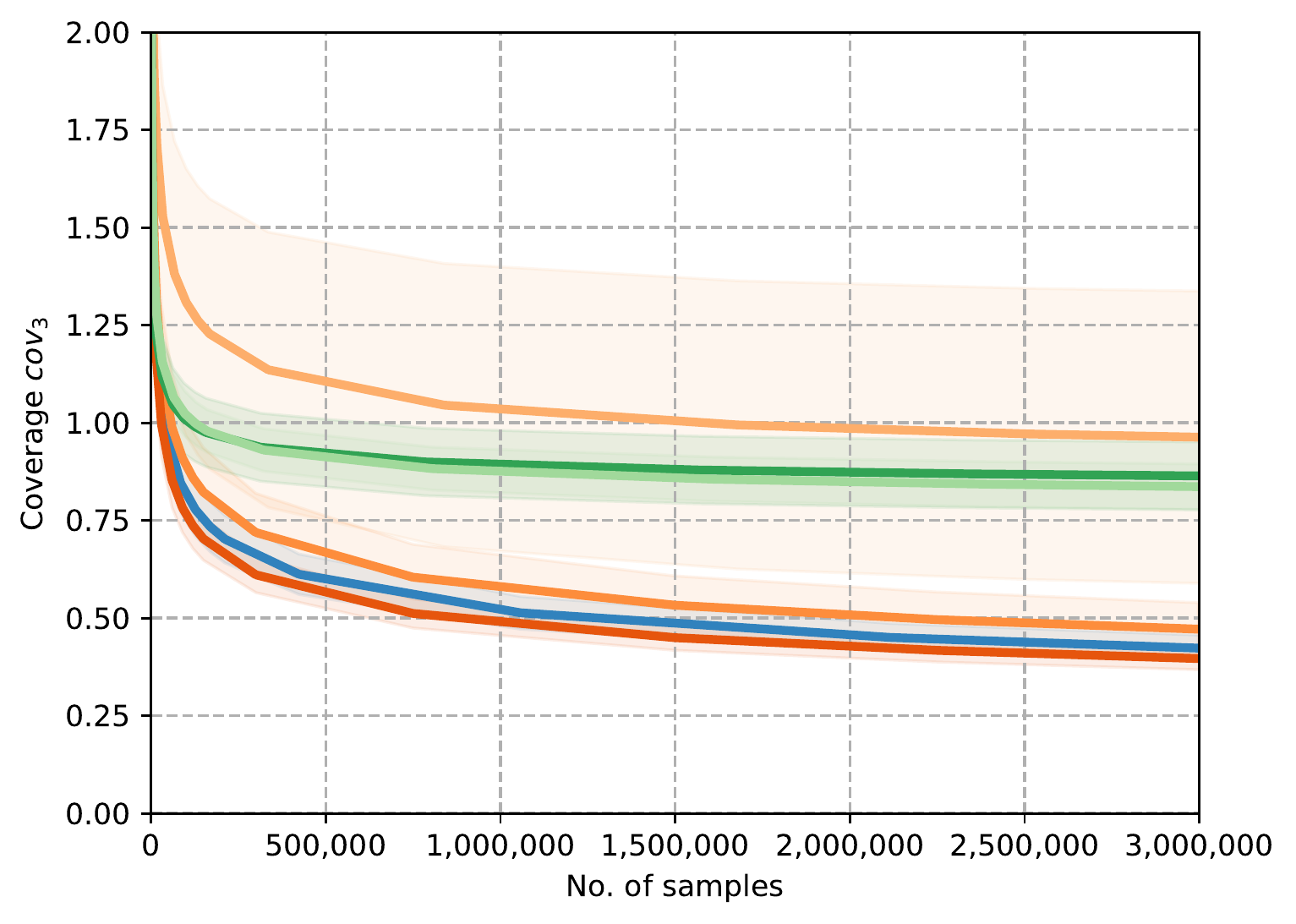}\\%
    \includegraphics[width=0.5\textwidth]{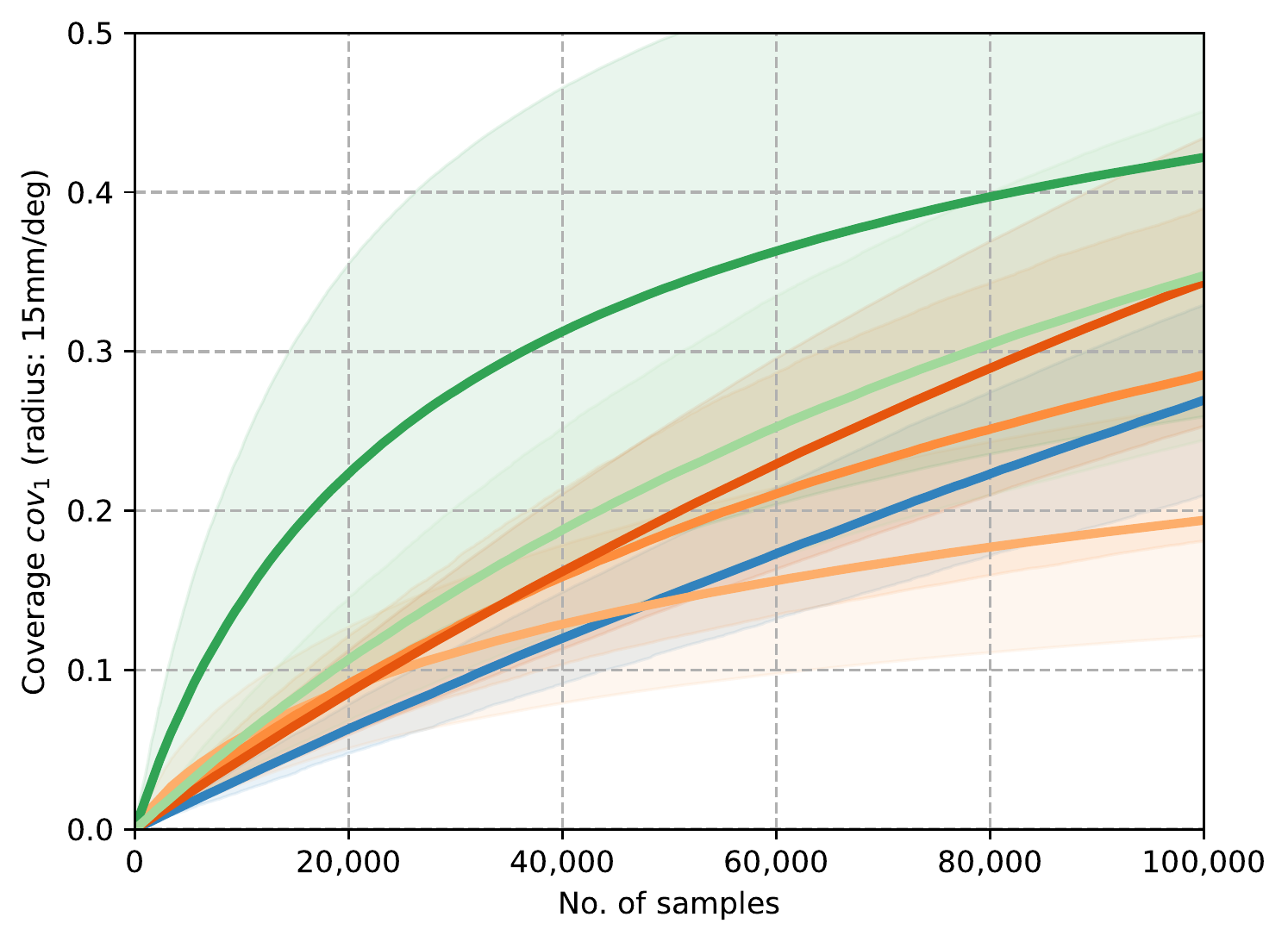}%
    \includegraphics[width=0.5\textwidth]{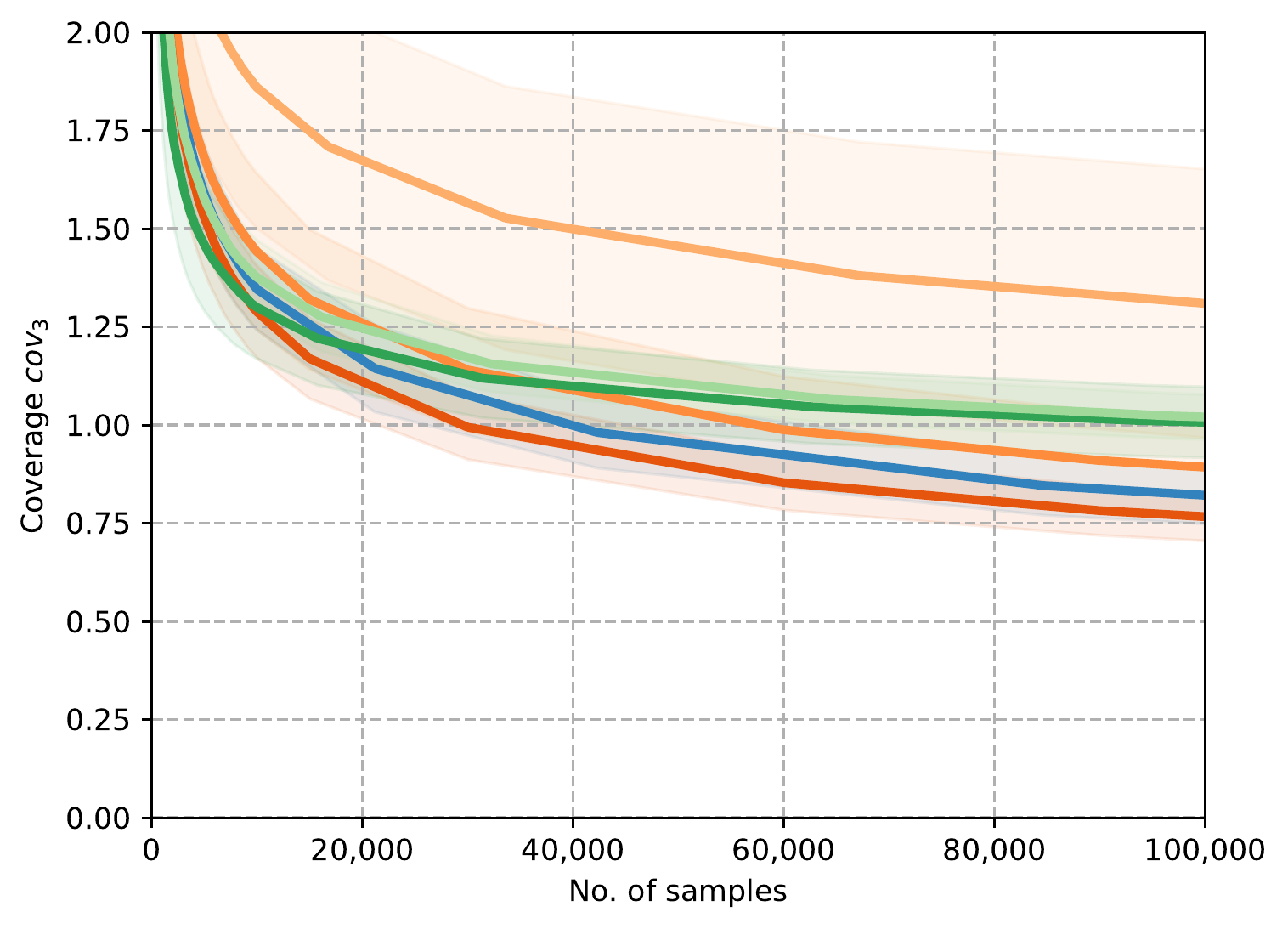}%
    \caption{Mean coverage and standard deviation over all objects for different sampling strategies. The lower plots magnify the curves during the first 100,000 samples.}
    \label{fig:coverage_avg_comparison}
\end{figure}

Fig.~\ref{fig:coverage_avg_comparison} shows a comparison of all grasp samplers averaged over all objects. We show curves for two coverage metrics~($\text{cov}_1$ and $\text{cov}_3$) for the first 3~million samples and a zoom-in on the first $100,000$~sampled grasps.

The uniform sampling scheme is the least biased one, attaining full coverage $\text{cov}_1$ within the first 3~million samples over all objects. The approach-based sampling strategies have a wide performance range depending on their parameterization. The surface(0, 0) strategy is the worst, it only samples grasps along the surface normals of objects. This leads to uncovered holes in the resulting grasp set, especially close to discontinuous structures such as edges. As a result e.g. the blades of the scissors cannot be grasped from all sides equally.
The surface($\frac{\pi}{2}, 0$) strategy does not suffer from this problem since it samples approach directions from a cone centered around the surface normals. Consequently, it is the second best sampling strategy in terms of coverage.
Including the same amount of variation when choosing the gripper's approach vector does not lead to high coverage, as shown by the curve of surface($0, \frac{\pi}{2}$).

The antipodal-based strategies perform not as good as the best approach-based strategy. Both of them saturate at around \SI{68}{\%}/\SI{75}{\%} coverage. Their bias is visualized in Fig.~\ref{fig:antipodal missing grasps}, where the reference grasps are shown that are farthest away from the ones sampled by~antipodal($\frac{\pi}{6}$). It can be seen, that grasping the lip of the meat can is not covered.

The lower plots in Fig.~\ref{fig:coverage_avg_comparison} show a magnifying view of the coverage performance during the first 100,000 samples. This is important if only a limited sampling budget is available. In this case the antipodal schemes, especially the antipodal($\frac{\pi}{6}$) scheme is the best one. Its exploitative behavior finds suitable grasps quicker than any other sampling strategy.

\subsubsection{Qualitative Grasp Differences}
\begin{figure}[t]
    \centering
    \includegraphics[width=0.25\textwidth,height=3.7cm,keepaspectratio]{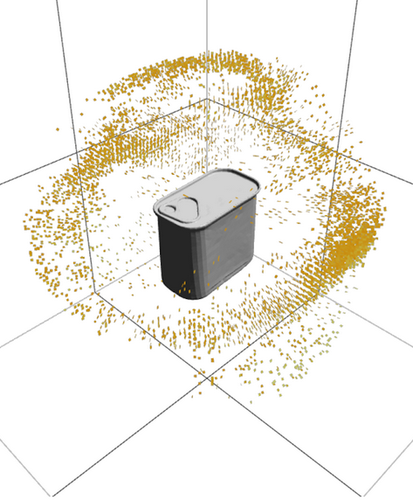}%
    \includegraphics[width=0.25\textwidth,height=3.7cm,keepaspectratio]{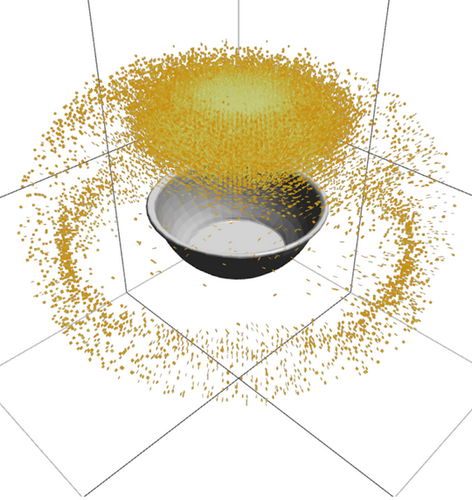}%
    \includegraphics[width=0.25\textwidth,height=3.7cm,keepaspectratio]{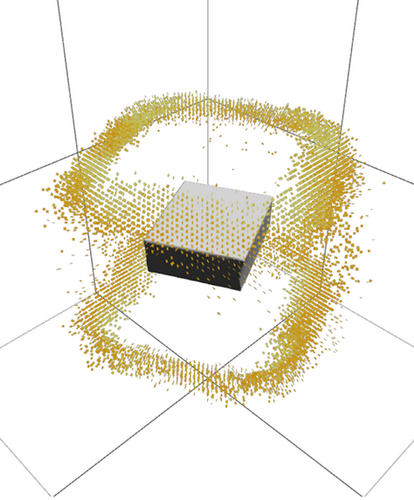}%
    \includegraphics[width=0.25\textwidth,height=3.7cm,keepaspectratio]{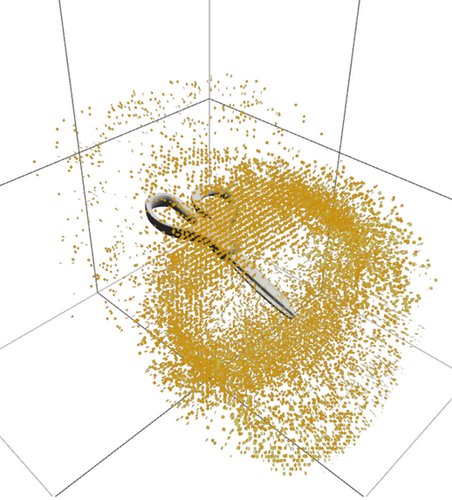}%
    \caption{Successful grasps of the potted meat can, bowl, gelatin box, and scissors that are missed by the antipodal-based sampling strategy.}
    \label{fig:antipodal missing grasps}
\end{figure}
The previous experiment showed that sampling heuristics that are more exploitative suffer from a high bias, i.e., they do not cover all possible grasps.
But what kind of grasps are missed?
To answer this question we computed the shortest distance for each successful reference grasp to the sampled grasps.
Fig.~\ref{fig:antipodal missing grasps} shows the most distant reference grasps for the antipodal($\frac{\pi}{6}$) scheme for various objects.
It can be seen that the antipodal sampler misses small-scale features such as the rim of the potted meat can. It also ignores approaches directed towards edges that result in successful grasps like shown with the gelatin box, bowl, and scissor blade.
See~\url{https://bit.ly/2HWEI2r} for more examples.

\subsubsection{Robust Grasp Coverage}
\begin{figure}[t]
    \centering
    \includegraphics[width=0.7\textwidth]{figures/coverage/legend2-crop.pdf}\\%
    \includegraphics[width=0.5\textwidth]{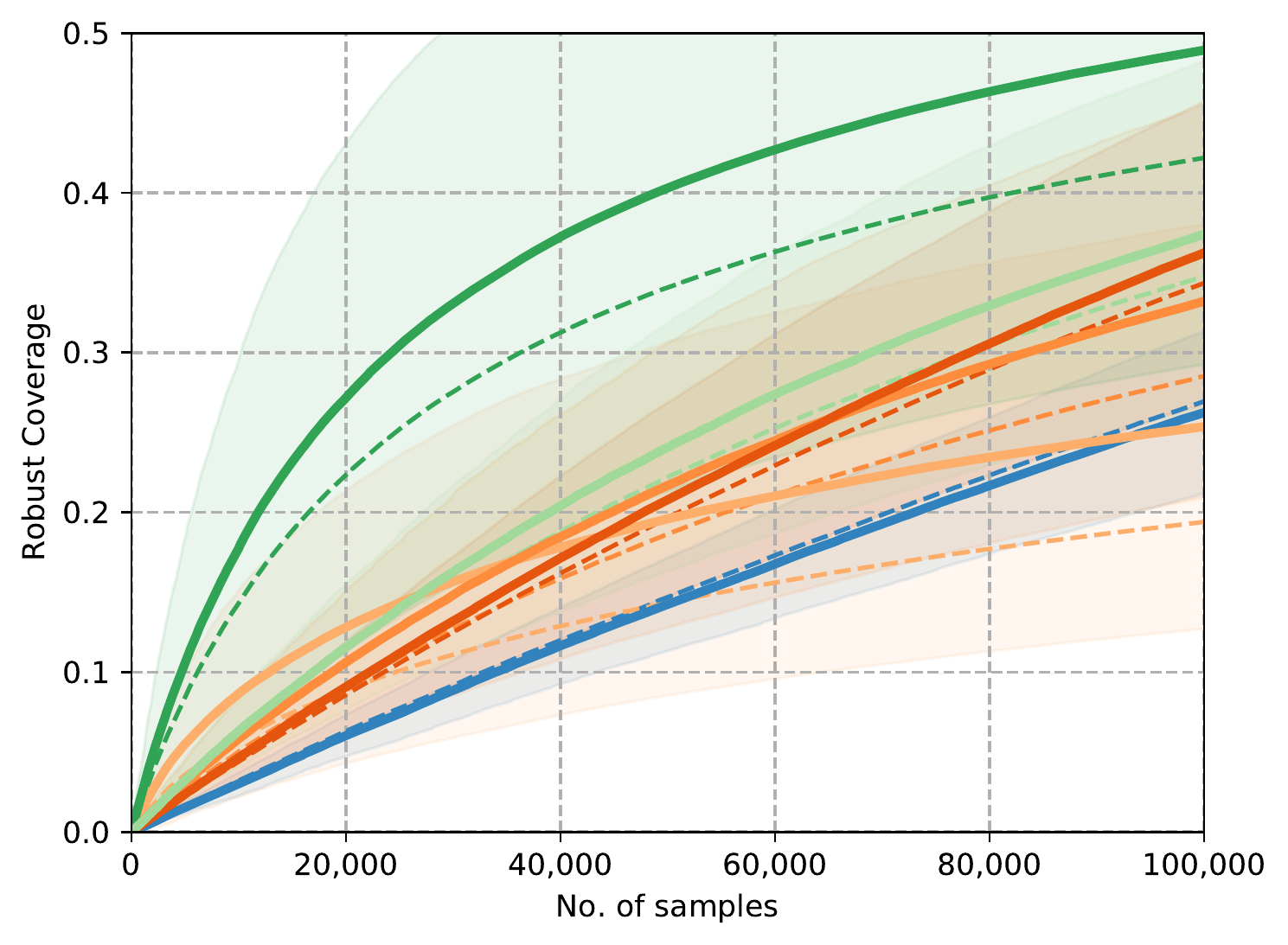}%
    \includegraphics[width=0.5\textwidth]{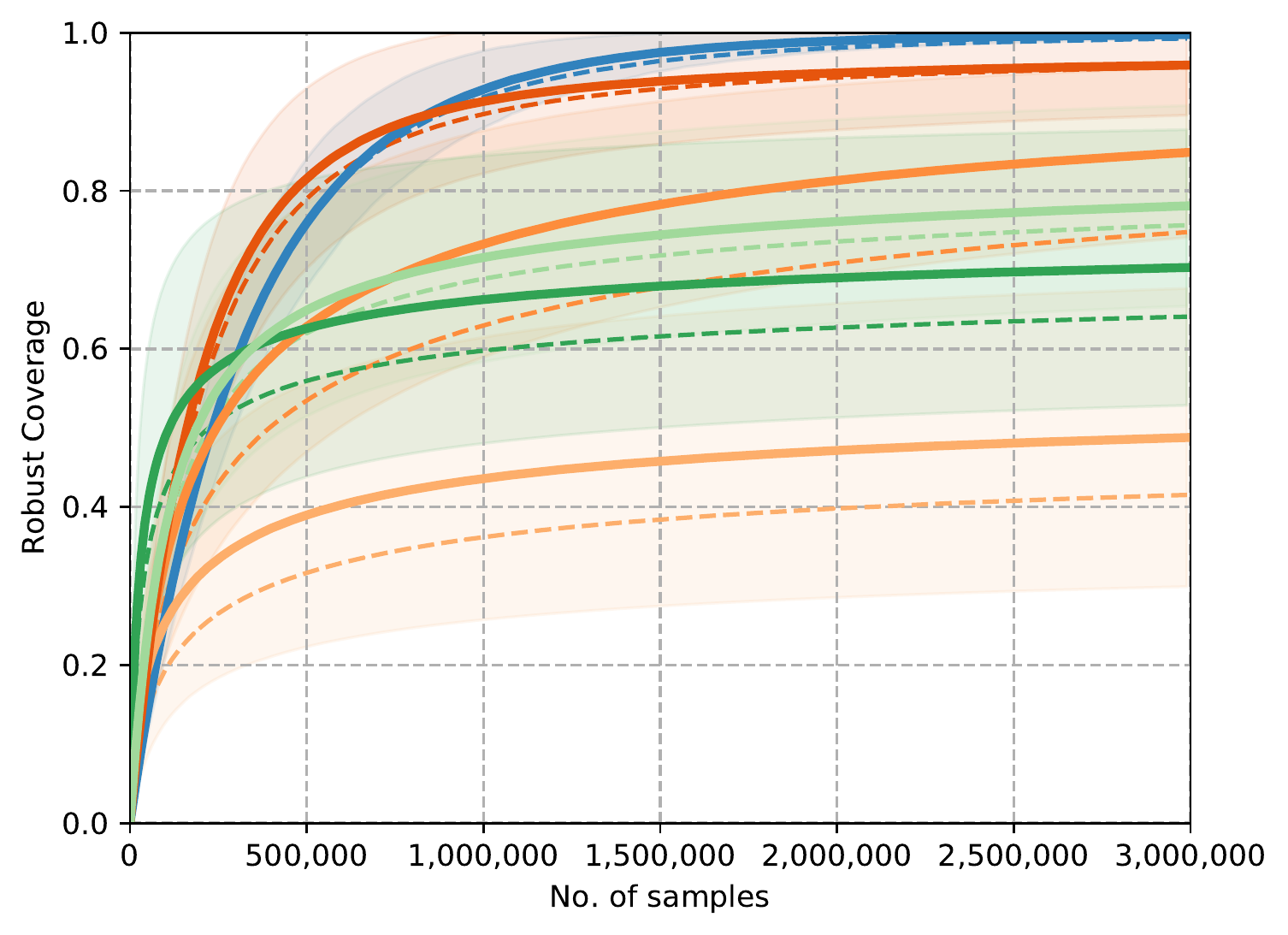}%
    \caption{Coverage for robust grasps for each sampling scheme. The dashed lines show the coverage on the original set~(Fig.~\ref{fig:coverage_avg_comparison}).}
    \label{fig:robust_grasp_coverage}
\end{figure}

We evaluated the grasp samplers also w.r.t. the set of robust grasps for each object as defined in Sec.~\ref{subsec:robust grasps}.
The results shown Fig.~\ref{fig:robust_grasp_coverage} reveal that the ranking of the different heuristics does not change.
Still some samplers focus more on robust grasps than others. While with fewer samples the antipodal($\frac{\pi}{6}$) scheme seems to gain the most coverage, asymptotically the approach($0, \pi$) scheme benefits the most by only considering robust grasps.

\subsubsection{Precision}
In a final experiment we compare the precision of different sampling schemes, i.e., the probability of a sampled grasp to be successful. 
The results in Table~\ref{tab:precision} show that a larger bias not necessarily leads to higher precision.
The uniform and approach($0, \pi$) strategy exhibit the lowest precision. The antipodal($\frac{\pi}{6}$) scheme has the highest precision. For learning approaches having a balanced set might be advantageous.

\begin{table}[th]
    \centering
    \begin{tabular}{|c|c|c|c|c|c|}
    \hline%
        \multirow{2}{*}{Uniform} & \multicolumn{3}{c|}{Approach} & \multicolumn{2}{c|}{Antipodal} \\
        & $(0, \pi)$ & $(0, 0)$ & $(\frac{\pi}{2}, 0)$ & $(\frac{\pi}{6})$ & $(\frac{\pi}{2})$ \\\hline
        $0.26$ & $0.29$ & $0.49$ & $0.36$ & $0.75$ & $0.42$ \\
        ($\pm0.15$) & ($\pm0.17$) & ($\pm0.28$) & ($\pm0.23$) & ($\pm0.26$) & ($\pm0.24$) 
    \\\hline%
    \end{tabular}
    \caption{Average precision (STD) of the different grasp sampling schemes.}
    \label{tab:precision}
\end{table}

%% file: 08_discussion.tex
\section{Discussion}
Our comparison of different grasp sampling schemes exposes a kind of bias-variance dilemma. Less constrained samplers such as the uniform one will cover all successful grasps for all objects but do so at the expense of poor sample efficiency. On the other hand more constrained heuristics can be efficient but might not capture the entire subspace of possible grasps.
The empirical evaluation revealed that the antipodal scheme is initially much more effective at capturing large parts of the grasp subspace compared to the approach-based schemes. But one needs to be aware of the imposed bias.
Note that for a given fixed sampling budget it is advantageous to chose a set rather than a sequence since it will lead to lower dispersion.

Our analysis focuses on the behavior of sampling heuristics as a function of the number of samples.
It assumes that the computational complexity of drawing a valid sample is comparable between different heuristics.
Although it is significantly more difficult, a more faithful comparison should look at the value of different heuristics per unit of computation.

\subsubsection*{Limitations}
Note that the simulation data has a few limitations: Due to the discretization there are aliasing effects shown by asymmetric grasp sets for symmetric objects.
Additionally, we do not simulate gravity or any contact constraints with the environment.
We also did not vary the internal DOF of the gripper.
Adding all these dimensions would impede us from simulating all possible grasps in a reasonable amount of time.

%% file: 09_conclusions.tex
\section{Conclusions}
We presented a dense data set of parallel-jaw grasps for 21~objects from the YCB data set. The data set is annotated with the results from running a physics simulation for more than a billion grasps. We showed that the quality of the simulation is reasonable, by using a model-based robotic system and transferring the successful grasps to the real world.

The data allowed us to quantify empirically for the first time the bias exposed by existing grasp sampling schemes. This will improve the understanding of data generation  for 6-DOF grasp learning algorithms. Fully capturing the entire grasp distribution is important in order to plan grasps that are conditioned on task or environmental constraints and go beyond simple pick-and-place scenarios.

\subsubsection*{ACKNOWLEDGMENT}
We thank Miles Macklin, Viktor Makoviychuk, and Nuttapong Chentanez for support with FleX.